%% file: bmvc_final.tex
\documentclass{bmvc2k}

\usepackage{booktabs}
\usepackage{multirow}
\usepackage{makecell}
\usepackage{subfigure}
\usepackage[colorlinks,urlcolor=blue,citecolor=red,bookmarks=false]{hyperref}

\graphicspath{{./org_images/}{./openmax_MAV/}{./stingrays/}{./regular1s/}}

\renewcommand\cap[3]{\caption[#2]{\label{#1}\textsc{#2} \small{#3}}}
\newcommand\ic[2][1]{\includegraphics[width=#1\textwidth]{#2}}
\newcommand\etal[1]{\textit{et al.}~\cite{#1}}

\title{Adversarial Robustness: \\Softmax versus Openmax}

\addauthor{Andras Rozsa}{arozsa@vast.uccs.edu}{1}
\addauthor{Manuel G\"unther}{mgunther@vast.uccs.edu}{1}
\addauthor{Terrance E. Boult}{tboult@vast.uccs.edu}{1}

\addinstitution{
 Vision and Security Technology (VAST) Lab\\
 University of Colorado Colorado Springs\\
 Colorado Springs, CO, USA
}

\def\etaltext{et al\bmvaOneDot}

\runninghead{Rozsa \protect\etaltext}{Adversarial Robustness: Softmax versus Openmax}

\begin{document}

\maketitle

\begin{abstract}

Deep neural networks (DNNs) provide state-of-the-art results on various tasks and are widely used in real world applications.
However, it was discovered that machine learning models, including the best performing DNNs, suffer from a fundamental problem: they can unexpectedly and confidently misclassify examples formed by slightly perturbing otherwise correctly recognized inputs.
Various approaches have been developed for efficiently generating these so-called adversarial examples, but those mostly rely on ascending the gradient of loss.
In this paper, we introduce the novel logits optimized targeting system (LOTS) to directly manipulate deep features captured at the penultimate layer.
Using LOTS, we analyze and compare the adversarial robustness of DNNs using the traditional Softmax layer with Openmax, which was designed to provide open set recognition by defining classes derived from deep representations, and is claimed to be more robust to adversarial perturbations. 
We demonstrate that Openmax provides less vulnerable systems than Softmax to traditional attacks, however, we show that it can be equally susceptible to more sophisticated adversarial generation techniques that directly work on deep representations.

\end{abstract}


\input{introduction}
\input{relatedwork}
\input{approach}
\input{experiments}
\input{conclusion}

\vspace{5pt}
\subsubsection*{Acknowledgments}

This research is based upon work funded in part by NSF IIS-1320956 and in part by the Office of the Director of National Intelligence (ODNI), Intelligence Advanced Research Projects Activity (IARPA), via IARPA R\&D Contract No. 2014-14071600012. The views and conclusions contained herein are those of the authors and should not be interpreted as necessarily representing the official policies or endorsements, either expressed or implied, of the ODNI, IARPA, or the U.S. Government. The U.S. Government is authorized to reproduce and distribute reprints for Governmental purposes notwithstanding any copyright annotation thereon.

\newpage
\bibliography{bmvc_review}
\end{document}

%% file: introduction.tex
\section{Introduction}

In recent years, deep neural networks (DNNs) have demonstrated impressive performance improvements on a wide range of challenging machine learning tasks \cite{hinton2012deep,szegedy2015going,parkhi2015deep,he2015deep}.
Despite the excellent results, our understanding of these networks is in question following the revealed intriguing properties of machine learning models \cite{szegedy2013intriguing}.
Since DNNs are able to learn high-level feature embeddings -- allowing them to be successfully adapted to different problems -- and generalize well, the discovery of adversarial examples by Szegedy \etal{szegedy2013intriguing} was rather astonishing.
Due to the excellent generalization properties, DNNs were expected to be robust to moderate distortions to their inputs, however, even imperceptibly small adversarial perturbations yield altered classifications.
Closely related to the problem of adversarial vulnerability, Nguyen \etal{nguyen2014deep} demonstrated that DNNs can also confidently misclassify the so-called fooling examples that are totally unrecognizable to humans.

The advances in deep learning boosts various research areas.
DNNs optimized for specific tasks can learn representations that generalize well to other datasets \cite{taigman2014deepface,parkhi2015deep}, and the extracted generic descriptors can be successfully used for other problems \cite{sharif2014cnn,donahue2014decaf} as well.
Therefore deep features extracted from these networks are widely used in the literature -- e.g., for different visual recognition tasks in biometrics \cite{liu2015deep,taigman2014deepface,sun2014deep,parkhi2015deep} -- even where the application of an end-to-end classification network would not be applicable.

One interesting application of the extracted generic descriptors is to ``open'' the traditionally closed set nature of end-to-end classification DNNs.
While recognition in the real world is open set and a trained recognition system should ideally reject unknown or unseen classes, DNNs are forced to choose from one of the known classes for any given input -- even if humans do not consider them meaningful.
To address the problem, Bendale and Boult \cite{bendale2016towards} introduced a new network layer called Openmax, which estimates the probability of an input being from an unknown class and, thereby, provides an open set recognition solution for DNNs.
After evaluating open set DNNs built upon pre-trained networks from the Caffe Model Zoo\footnote{\scriptsize\url{https://github.com/BVLC/caffe/wiki/Model-Zoo}} on the ILSVRC-2012 validation dataset \cite{russakovsky2015imagenet}, the authors concluded that Openmax improves the robustness of those networks to adversarial and fooling images -- by either correcting or detecting those -- and their robustness significantly outperforms basic DNNs as well as DNNs using thresholds for Softmax probabilities.

In this paper, we introduce the logits optimized targeting system (LOTS) designed to perturb samples in such ways that their deep feature representations mimic any selected target activations at the penultimate layer, which are also called \textit{logits}.
We demonstrate the effectiveness of LOTS in terms of forming high quality adversarial and unrecognizable examples by targeting classes in two ways including mimicking the mean activation vectors (MAV) of classes Openmax uses for open set recognition.
We analyze and compare the robustness to adversarial and unrecognizable images of the BVLC-GoogLeNet
network using the traditional Softmax layer with the same network utilizing Openmax.
Our results show that Openmax outperforms the traditional Softmax layer when LOTS is used to directly target classes, however, it is equally vulnerable to adversarial and unrecognizable images when LOTS is utilized to mimic the class representations that Openmax relies on.

%% file: relatedwork.tex
\section{Related Work}

Since adversarial examples were discovered, researchers have proposed various approaches capable of reliably finding such perturbations.
While the first method of Szegedy \etal{szegedy2013intriguing} relied on the computationally expensive box-constrained optimization technique (L-BFGS), Goodfellow \etal{goodfellow2014explaining} introduced a more lightweight, yet very effective approach -- called the fast gradient sign (FGS) -- which relies on the sign of the gradient of loss with respect to the input.
Rozsa \etal{rozsa2016adversarial} later demonstrated that with a slight modification of FGS, namely, by dropping the use of the sign, the formalized fast gradient value (FGV) approach generates very different, high quality adversarial examples.

The aforementioned adversarial generation techniques simply ascend the gradient of training loss for the given input until the particular original class does not have the highest prediction probability.
Recently, Kurakin \etal{kurakin2016adversarial} proposed extensions over the simple FGS method in order to be able to target a specific class, or by calculating and applying gradients iteratively compared to using a single gradient via FGS.

Other approaches that do not rely on using the gradient of training loss were also proposed in the literature.
Rozsa \etal{rozsa2016adversarial} introduced the hot/cold approach utilizing a Euclidean loss with arbitrarily selected target classes on the pre-Softmax layer -- the so-called logits --  and uses its gradients as directions for forming adversarial perturbations.
While this approach does not use the gradient of the training loss, eventually, it still targets training classes and cannot be used to directly manipulate deep features.

Unlike the previous techniques, the approach introduced by Sabour \etal{sabour2015adversarial} forms adversarial examples that cause misclassifications by mimicking the internal representations of the targeted inputs.
While this method works on deep features and can potentially be used to attack systems that utilize deep representations extracted from DNNs, it relies on using the computationally expensive L-BFGS algorithm, which limits its practical application.

The formation of fooling examples, i.e., images that are unrecognizable to humans but DNNs confidently classify them as recognizable objects, is closely related to adversarial example generation.
In short, one can summarize the difference simply as ``painting to a non-blank canvas having a different background'' -- where the background is either recognizable or unrecognizable.
Nguyen \etal{nguyen2014deep} used evolutionary algorithms or gradient ascent for forming two types of fooling images:
indirectly encoded or regular images and directly encoded or irregular images.
As we can see in Fig.~\ref{org:i} and \subref{org:j}, regular images possess patterns, while irregular images shown in Fig.~\ref{org:k} and \subref{org:l} do not.

To provide open set recognition, Openmax \cite{bendale2016towards} estimates the probabilities of inputs being \textit{unknown} by adapting meta-recognition concepts to activation patterns of the penultimate layer.
While deeper features could also be utilized, Openmax working on logits was reported to greatly reduce the number of obvious classification errors.
Ideally, a DNN with Openmax detects unrecognizable images as \textit{unknown} and corrects adversarial images by classifying them as their original class or \textit{unknown}.
The authors tested with and reported their results on adversarial and fooling images generated on basic DNNs utilizing the Softmax layer.

Since Openmax is implemented external to DNNs and operates on extracted deep features, end-to-end adversarial generation techniques cannot be used for forming adversarial or unrecognizable examples on this layer.
Our novel LOTS method is capable of directly manipulating the deep representations of images that Openmax relies on and, therefore, we can directly attack DNNs using either the Softmax or Openmax layer.
LOTS can be extended to deeper layers as well, and it shows similarities to the technique of Sabour \etal{sabour2015adversarial} in terms of directly adjusting deep features -- without using L-BFGS.

%% file: approach.tex
\section{Approach}

This section describes the deep neural networks (DNNs) we experiment with, introduces our novel approach to form adversarial and unrecognizable images on those systems and, finally, presents the metric that we use for quantifying the quality of the generated examples.

\subsection{Targeted DNNs: Closed Set vs. Open Set Recognition}

For our experiments, we use the publicly available BVLC-GoogLeNet\footnote{\scriptsize\url{https://github.com/BVLC/caffe/tree/master/models/bvlc_googlenet}} network, which is the Berkeley-trained version of the DNN designed by Szegedy \etal{szegedy2015going}, optimized on the ILSVRC-2012 dataset \cite{russakovsky2015imagenet}.
To compare the robustness of the traditional network with an open set DNN, we have replaced the Softmax layer with our implementation of Openmax.

Openmax works on extracted deep features of the penultimate layer, which are also called activation vectors (AVs).
First, we have extracted AVs from the $224\times224$-pixel center crops of the correctly classified training images that were rescaled to $256\times256$ pixels.
Second, from the extracted activation vectors we have calculated a single point to represent each known class by using their mean activation vector (MAV).
We use Openmax parameters as defined in \cite{bendale2016towards}, and we verified our implementation with the authors.
Finally, given an input image and the MAVs, using the activation vector extracted from the input Openmax calculates probabilities for known classes and for being \textit{unknown}.
Similarly to Softmax, Openmax normalizes probabilities to sum up to one.
To determine the classification of the open set DNN, we simply take the class with the highest prediction probability.

\subsection{LOTS - Logits Optimized Targeting System}
\label{sec:lots}

Let us consider a general deep neural network (DNN) $f$ with weights $w$ in a layered structure, i.e., having layers $y^{(l)}, l = \{1, \ldots,L\}$, with their corresponding weights $w^{(l)}$.
For a given input $x$, the output of the DNN can be formalized as:
\begin{equation}
  f(x) = y^{(L)}\left(y^{(L-1)}\left({\ldots\left({y^{(1)}(x)}\right)\ldots}\right)\right),
\end{equation}
while the deep feature of input $x$ at layer $l$ is:
\begin{equation}
  f^{(l)}(x) = y^{(l)}\left(y^{(l-1)}\left(\ldots\left(y^{(1)}(x)\right)\ldots\right)\right).
\end{equation}

Our logits optimized targeting system (LOTS) adjusts the activation vector (AV) of input $x_o$ extracted from the penultimate layer $L-1$ (so-called logits) to get closer to a target $t$.
In order to do so, we use a Euclidean loss defined on the activation vector $f^{(L-1)}(x_o)$ of input $x_o$ and the target $t$, and apply its gradient with respect to input $x_o$, formally:
\begin{equation}
\label{lots}
  \eta^{(L-1)}(x_o,t) = {{\nabla}_{x_o}} \left( \frac{1}{2} \left\| t - f^{(L-1)}(x_o) \right\|^2 \right).
\end{equation}

The target $t$ can be chosen without any constraints.
Furthermore, LOTS can be applied to deep features of any layer.
We use the activation vectors of the penultimate layer with a Euclidean loss as Openmax works on logits in a Euclidean space.

To produce high quality examples with less perceptible perturbations, we use LOTS iteratively.
Our algorithm utilizes a scaled gradient with L$_\infty=1$ to move faster to the targeted activation vector until the perturbed image is classified as desired.
Iterative LOTS forms image $x_p$ with discrete pixel values in $\left[0,255\right]$, however, while taking steps towards the target, it temporarily utilizes $x_p'$ with non-discrete pixel values in order to obtain better adversarial quality.
This ``step-and-adjust'' algorithm extracts the activation vector (AV) of the temporary image $x_p'$ (initialized to $x_o$), calculates the gradient as defined by Eq.~\eqref{lots}, and uses that to take a step to get closer to the target $t$.
These steps are repeated with newly extracted AVs until the DNN classifies $x_p$ as the targeted class.

\subsection{Adversarial Quality}

To compare the robustness of closed and open set DNNs to adversarial and unrecognizable images, we need to assess the quality of the produced examples.
Although L$_p$ norms are commonly used to quantify perturbations, researchers~\cite{sabour2015adversarial,rozsa2016adversarial} concluded that those measures do not match well to human perception.
Therefore, we use the psychometric called the perceptual adversarial similarity score (PASS) introduced by Rozsa \etal{rozsa2016adversarial}.
While L$_p$ norms focus strictly on the perturbation -- regardless of its visibility on the distorted image -- PASS is designed to quantify the similarity of the original image $x_o$ and the perturbed image $x_p$ with respect to human perception.

To calculate PASS, the image pair is first aligned by maximizing the enhanced correlation coefficient (ECC)~\cite{evangelidis2008parametric} with homography transform $\Psi(x_p, x_o)$, and then their structural similarity is quantified by using the structural similarity (SSIM) index~\cite{wang2004image}.
The alignment step via ECC takes place before SSIM calculation as small translations or rotations can remain imperceptible to the human eye and, therefore, PASS eliminates those before the structural similarity of the image pair is calculated.
In summary, PASS can be formalized as follows:
\vspace{-2pt}
\begin{equation}
\label{pass}
\textup{PASS}\left( x_p, x_o \right) = \textup{SSIM}\left( \Psi \left( x_p,x_o \right),\, x_o \right),
\end{equation}
where $\textup{PASS}(x_p, x_o)=1$ defines perfect similarity.

Since the structural similarity (SSIM) index can only be calculated on grayscale images, we align the converted grayscale images using OpenCV's ECC with termination criteria of 100 iterations or $\epsilon=0.01$, and then we calculate the perceptual adversarial similarity score (PASS) of the aligned images using SSIM.\footnote{\scriptsize\url{http://isit.u-clermont1.fr/~anvacava/codes/ssim.py}}

%% file: experiments.tex
\section{Experiments}

The ultimate goal of this paper is to answer the question whether open set DNNs are less susceptible to adversarial and unrecognizable images than the traditional closed set DNNs.
This section describes our experiments and presents our results.

\begin{figure*}[t]
  \centering
  
 \subfigure[][\label{org:a}Bee\,\,(0.155)]{\ic[.24]{bee}} \hspace{-2pt}
 \subfigure[][\label{org:b}Guitar\,\,(0.837)]{\ic[.24]{guitar}} \hspace{-2pt}
 \subfigure[][\label{org:c}Hummingbird\,\,(0.876)]{\ic[.24]{hummingbird}} \hspace{-2pt}
 \subfigure[][\label{org:d}Marmot\,\,(0.393)]{\ic[.24]{marmot}} \\
 \vspace{-4pt}
 
 \subfigure[][\label{org:e}Spider Web\,\,(0.528)]{\ic[.24]{spider_web}} \hspace{-2pt}
 \subfigure[][\label{org:f}Stingray\,\,(0.922)]{\ic[.24]{stingray}} \hspace{-2pt}
 \subfigure[][\label{org:g}Submarine\,\,(0.999)]{\ic[.24]{submarine}} \hspace{-2pt}
 \subfigure[][\label{org:h}Wine Bottle\,\,(0.787)]{\ic[.24]{wine_bottle}} \\
 \vspace{-4pt}
 
 \subfigure[][\label{org:i}Regular 1\,\,(1.000)]{\ic[.24]{regular1}} \hspace{-2pt}
 \subfigure[][\label{org:j}Regular 2\,\,(0.947)]{\ic[.24]{regular2}} \hspace{-2pt}
 \subfigure[][\label{org:k}Irregular 1\,\,(0.468)]{\ic[.24]{irregular1}} \hspace{-2pt}
 \subfigure[][\label{org:l}Irregular 2\,\,(0.846)]{\ic[.24]{irregular2}} 

\cap{fig:org}{Known, Regular, and Irregular Images.}{Images in \subref{org:a}-\subref{org:h} are correctly classified by BVLC-GoogLeNet networks with both Softmax and Openmax. While Softmax classifies regular (\subref{org:i} \textit{coral reef}, \subref{org:j} \textit{coil}) and irregular (\subref{org:k} \textit{velvet}, \subref{org:l} \textit{window screen}) images as known classes, Openmax labels them as \textit{unknown}. Sub-captions contain the prediction certainty of Openmax.}
\end{figure*}

\subsection{Attacks on Softmax and Openmax}

\newcommand\rt[1]{\textcolor{red}{(#1)}}

\begin{figure*}[t]
  \centering
  
 \subfigure[][\label{adv:a}PASS{\,=\,}0.997\,\,\rt{0.100}]{\ic[.24]{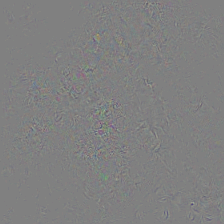}} \hspace{-2pt}
 \subfigure[][\label{adv:b}PASS{\,=\,}0.996\,\,\rt{0.062}]{\ic[.24]{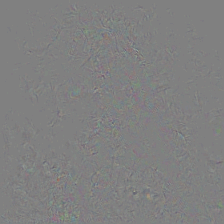}} \hspace{-2pt}
 \subfigure[][\label{adv:c}PASS{\,=\,}0.997\,\,\rt{0.159}]{\ic[.24]{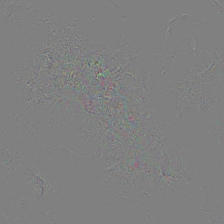}} \hspace{-2pt}
 \subfigure[][\label{adv:d}PASS{\,=\,}0.996\,\,\rt{0.330}]{\ic[.24]{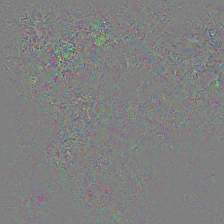}} \\
 \vspace{-4pt}
 
 \subfigure[][\label{adv:e}PASS{\,=\,}0.997\,\,\rt{0.244}]{\ic[.24]{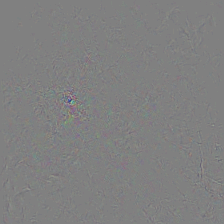}} \hspace{-2pt}
 \subfigure[][\label{adv:f}PASS{\,=\,}0.998\,\,\rt{0.289}]{\ic[.24]{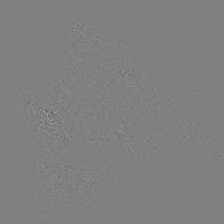}} \hspace{-2pt}
 \subfigure[][\label{adv:g}PASS{\,=\,}0.982\,\,\rt{0.118}]{\ic[.24]{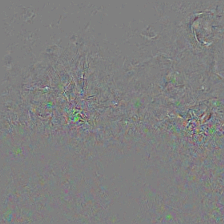}} \hspace{-2pt}
 \subfigure[][\label{adv:h}PASS{\,=\,}0.994\,\,\rt{0.215}]{\ic[.24]{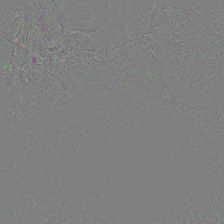}} \\
 \vspace{-4pt}
 
 \subfigure[][\label{adv:i}PASS{\,=\,}0.975\,\,\rt{0.367}]{\ic[.24]{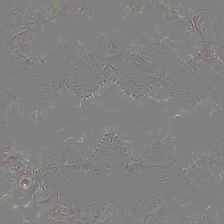}} \hspace{-2pt}
 \subfigure[][\label{adv:j}PASS{\,=\,}0.992\,\,\rt{0.077}]{\ic[.24]{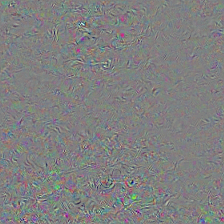}} \hspace{-2pt}
 \subfigure[][\label{adv:k}PASS{\,=\,}0.999\,\,\rt{0.156}]{\ic[.24]{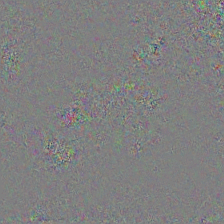}} \hspace{-2pt}
 \subfigure[][\label{adv:l}PASS{\,=\,}0.999\,\,\rt{0.173}]{\ic[.24]{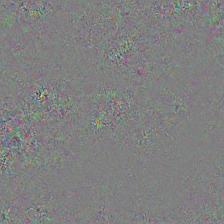}} 

\cap{fig:adv}{Perturbations on Openmax.}{These perturbations make Fig.~\ref{fig:org} images mimic the mean activation vector (MAV) of the goldfish class and, thus, they are classified as \textit{goldfish}. Sub-captions contain the PASS between original and perturbed images, and the prediction certainty of \textit{goldfish} via Openmax. For better visualization, perturbations are magnified by a factor of 5.}
\end{figure*}

To assess the capabilities of LOTS and compare the robustness of closed and open set DNNs, we use a dozen images to generate adversarial or unrecognizable examples.
While we intended to purely focus on analyzing the robustness of these DNNs to adversarial perturbations, experimenting with unrecognizable images can answer the question whether Openmax is capable of detecting such samples as \textit{unknown}. 
We have selected eight $224\times224$-pixel center crops from the rescaled $256\times256$-pixel images of the ILSVRC-2012 validation dataset that are correctly classified by both the closed and open set DNNs.
These images of known training classes are displayed in Fig.~\ref{org:a}-\subref{org:h}.
Throughout our experiments, we aim to generate adversarial examples from these images that are classified incorrectly as other possible known classes.
Considering the 1,000 classes, this yields 999 possible adversarial examples for each image.
To produce unrecognizable images, we use iterative LOTS simply to ``paint on non-blank canvasses that contain unrecognizable objects.''
Therefore, we have two regular images generated by Nguyen \etal{nguyen2014deep} shown in Fig.~\ref{org:i} and \subref{org:j} that we rescaled to $224\times224$ pixels.
Due to the combination of rescaling and using a different DNN, these examples are classified differently than in \cite{nguyen2014deep}.
To obtain the irregular images displayed in Fig.~\ref{org:k} and \subref{org:l}, we generated two $224\times224$-pixel images containing uniformly distributed random noise in $\left[0,255\right]$.
As expected, Openmax providing open set recognition classifies these regular and irregular images as \textit{unknown}.
Considering the classifications of regular and irregular images, there are 999 and 1,000 known classes to aim at when we form unrecognizable images on the closed and open set DNNs, respectively.

We conduct four sets of experiments using the iterative logits optimized targeting system (LOTS) approach (cf. Sec. \ref{sec:lots}).
\emph{First}, on the closed set DNN we use iterative LOTS on the extracted activation vectors to target known classes.
Since LOTS operates on logits, we define target $t$ at the logits layer that yields one-hot vector in the Softmax layer.
Particularly, to target a specific class we form a class aiming vector (CAV) $t$ with a value of $100$ for the targeted class and $-100$ for others.
This is the traditional way for forming adversarial perturbations on end-to-end classification networks.
\emph{Second}, we use iterative LOTS on the closed set DNN for forming adversarial and unrecognizable images by making their AVs mimic the targeted mean activation vectors (MAVs).
Compared to the first approach, LOTS still works on deep features of images extracted from the same layer, however, this time those AVs are manipulated to get closer to the MAVs that Openmax relies on.  
 
\emph{Third}, we use iterative LOTS on the open set DNN to form adversarial and unrecognizable examples with AVs  targeting the class aiming vectors (CAVs) as described in the first approach.
Out of our four sets of experiments, this is the closest to the one that Bendale and Boult~\cite{bendale2016towards} ran to assess the robustness of Openmax to adversarial and unrecognizable images.
We generate examples on this network, namely, images classified by Openmax as their targeted classes.
\emph{Fourth}, on the open set DNN we use iterative LOTS to form adversarial and unrecognizable examples with their AVs mimicking the mean activation vectors (MAVs) that Openmax uses as class representations.

To limit the computational costs, iterative LOTS is restricted to 500 steps.
In case this limit is exceeded for a given image, the attempt is considered a failure.
Otherwise, if the perturbed example is classified as the targeted class, LOTS stops in that particular step -- regardless what the prediction certainty of the network is.
This way iterative LOTS produces adversarial and unrecognizable images that contain the smallest possible perturbations yielding the desired classifications. 

\begin{figure*}[!ht]
  \centering
  
 \subfigure[][\label{types:a}\textsc{SM/CAV:}\,\,PASS{\,=\,}0.997\,\,\rt{0.282}]{\ic[.24]{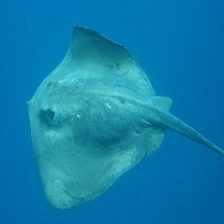}\hspace{2pt}\ic[.24]{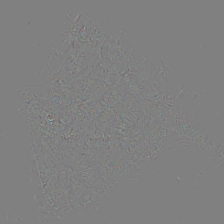}} \hspace{0pt}
 \subfigure[][\label{types:b}\textsc{SM/MAV:}\,\,PASS{\,=\,}0.998\,\,\rt{0.269}]{\ic[.24]{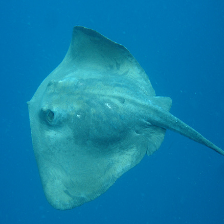}\hspace{2pt}\ic[.24]{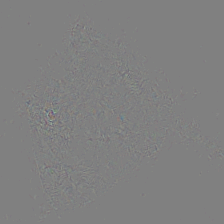}} \\
 \vspace{-4pt}
 
 \subfigure[][\label{types:c}\textsc{OM/CAV:}\,\,PASS{\,=\,}0.997\,\,\rt{0.230}]{\ic[.24]{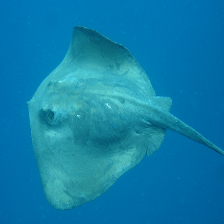}\hspace{2pt}\ic[.24]{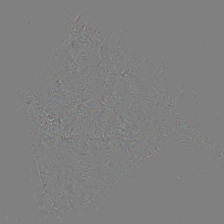}} \hspace{0pt}
 \subfigure[][\label{types:d}\textsc{OM/MAV:}\,\,PASS{\,=\,}0.998\,\,\rt{0.289}]{\ic[.24]{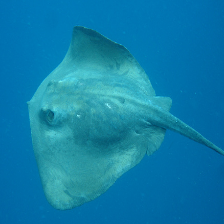}\hspace{2pt}\ic[.24]{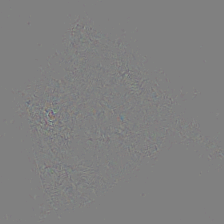}}\\
 \vspace{-4pt}

 \subfigure[][\label{types:e}\textsc{SM/CAV:}\,\,PASS{\,=\,}0.972\,\,\rt{0.311}]{\ic[.24]{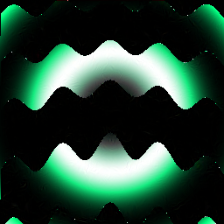}\hspace{2pt}\ic[.24]{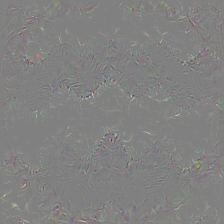}} \hspace{0pt}
 \subfigure[][\label{types:f}\textsc{SM/MAV:}\,\,PASS{\,=\,}0.972\,\,\rt{0.408}]{\ic[.24]{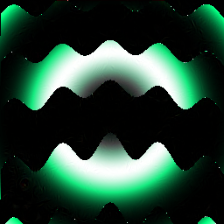}\hspace{2pt}\ic[.24]{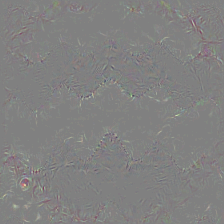}}\\
 \vspace{-4pt}
 
 \subfigure[][\label{types:g}\textsc{OM/CAV:}\,\,PASS{\,=\,}0.962\,\,\rt{0.245}]{\ic[.24]{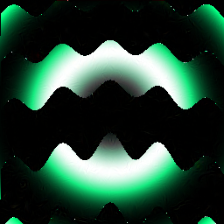}\hspace{2pt}\ic[.24]{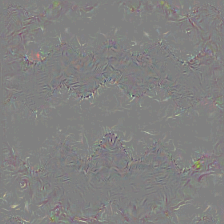}} \hspace{0pt}
 \subfigure[][\label{types:h}\textsc{OM/MAV:}\,\,PASS{\,=\,}0.975\,\,\rt{0.367}]{\ic[.24]{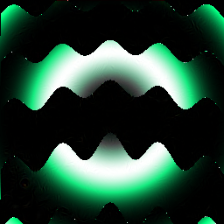}\hspace{2pt}\ic[.24]{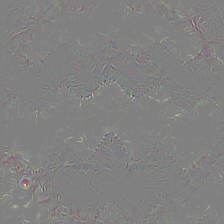}}

\cap{fig:types}{Softmax vs.~Openmax: Adversarial and Unrecognizable Images.}{This figure shows adversarial and regular unrecognizable examples paired with their corresponding perturbations yielding misclassifications as \textit{goldfish} on BVLC-GoogLeNet networks with either Softmax (SM) or Openmax (OM). Examples are formed by targeting the class aiming vector (CAV) or the mean activation vector (MAV) of the targeted class. Sub-captions contain the PASS between original and perturbed images, and the prediction certainty of \textit{goldfish} via Softmax or Openmax, respectively. For better visualization, perturbations are magnified by a factor of 5.}
\end{figure*}

\subsection{Results}

The results obtained by conducting the four sets of experiments using the selected images of Fig.~\ref{fig:org} are presented in Tab.~\ref{tab:results}.
Comparing the collected metrics on the two types of attacks on the closed set DNN with the Softmax layer, we can conclude that iterative LOTS targeting the class aiming vector (CAV) produces examples with very similar quality to the other type of attack where activation vectors mimic mean activation vectors (MAVs) of classes.
However, with the former approach LOTS is able to reach the targeted classes more frequently while forming adversarial and unrecognizable images.

On the open set DNN, iterative LOTS is less effective with CAV targets.
We can observe that while, in general, the quality of the produced examples is maintained at a similarly high level as on the closed set DNN, a large proportion of the targeted classes cannot be reached.
The proportion of successful attacks varies among the different images, the lowest rate is measured on the marmot shown in Fig.~\ref{org:d} with $64.5\,\%$.
Contrarily, when iterative LOTS is used with MAV-targets, Openmax is as vulnerable to adversarial and unrecognizable images as the closed set DNN with the Softmax layer.
This is highlighted by both the quality and the quantity of the formed examples.
To demonstrate how easily the images of known classes, as well as regular and irregular examples can be turned to a particular known class on the open set DNN, we show perturbations for each in Fig.~\ref{fig:adv} that turn images of Fig.~\ref{fig:org} to be classified as \textit{goldfish}, respectively.
For example, the activation vector of a confidently ($92.2\,\%$) and correctly classified \textit{stingray} shown in Fig.~\ref{org:f} can be manipulated by the perturbation displayed in Fig.~\ref{adv:f} to achieve the desired classification as \textit{goldfish}, with $28.9\,\%$ confidence.
The perturbed example is displayed in Fig.~\ref{types:d} -- it is a high quality adversarial example formed by an imperceptibly small perturbation as indicated by the high PASS.
Please note that we magnified those perturbations for better visualization by a factor of 5.

\begin{table*}[t]
\vspace{5pt}
\cap{tab:results}{Adversarial and Unrecognizable Images.}{These results are obtained using iterative LOTS with the listed ImageNet samples as well as regular and irregular images. With each, we attacked all 999 or 1,000 possible classes by targeting the class aiming vector (CAV) and the mean activation vector (MAV) of the class. We list the mean and standard-deviation of PASS and the percentage of successful attacks when perturbed images were classified as their targeted class.}
\vspace{5pt}
\centering
\scriptsize
\setlength{\tabcolsep}{3pt}
\resizebox{.99\textwidth}{!}{%
\begin{tabular}{c|c@{\hspace{2pt}}rc@{\hspace{2pt}}r|c@{\hspace{2pt}}rc@{\hspace{2pt}}r}
  \toprule
  \multirow{2}{1cm}{\centering\textsc{Original Image}}	& \multicolumn{4}{c|}{\textsc{Softmax Classification}} & \multicolumn{4}{c}{\textsc{Openmax Classification}} \\
    	& \multicolumn{2}{c}{\textsc{CAV Target}} & \multicolumn{2}{c|}{\textsc{MAV Target}} & \multicolumn{2}{c}{\textsc{CAV Target}} & \multicolumn{2}{c}{\textsc{MAV Target}} \\
  \midrule
  bee			& $0.995\pm0.002$ & $(100.0\%)$	& $0.993\pm0.004$ & $(100.0\%)$ 	& $0.990\pm0.006$ & $(82.3\%)$ & $0.992\pm0.004$ 	& $(100.0\%)$\\
  guitar 		& $0.997\pm0.001$ & $(100.0\%)$ 	& $0.996\pm0.002$ & $(100.0\%)$ 	& $0.996\pm0.003$ & $(89.2\%)$ & $0.996\pm0.002$ 	& $(100.0\%)$\\
  hummingbird 	& $0.994\pm0.004$ & $(100.0\%)$ 	& $0.992\pm0.005$ & $(99.9\%)$	& $0.988\pm0.008$ & $(80.8\%)$ & $0.992\pm0.005$ 	& $(99.7\%)$\\
  marmot 		& $0.998\pm0.001$ & $(100.0\%)$ 	& $0.997\pm0.002$ & $(99.3\%)$ 	& $0.996\pm0.003$ & $(64.5\%)$ & $0.997\pm0.002$ 	& $(99.5\%)$\\
  spider web 	& $0.993\pm0.003$ & $(100.0\%)$ 	& $0.991\pm0.005$ & $(99.9\%)$ 	& $0.988\pm0.007$ & $(82.0\%)$ & $0.991\pm0.005$ 	& $(99.9\%)$\\
  stingray 		& $0.995\pm0.002$ & $(100.0\%)$ 	& $0.993\pm0.005$ & $(99.7\%)$ 	& $0.991\pm0.007$ & $(83.4\%)$ & $0.993\pm0.005$ 	& $(99.8\%)$\\
  submarine		& $0.992\pm0.003$ & $(99.3\%)$ 	& $0.991\pm0.004$ & $(96.9\%)$ 	& $0.990\pm0.005$ & $(76.7\%)$ & $0.991\pm0.004$ 	& $(97.2\%)$\\
  wine bottle 	& $0.996\pm0.002$ & $(100.0\%)$	& $0.996\pm0.002$ & $(99.8\%)$ 	& $0.995\pm0.003$ & $(91.0\%)$ & $0.996\pm0.002$ 	& $(99.8\%)$\\
  regular 1 		& $0.960\pm0.016$ & $(99.9\%)$ 	& $0.952\pm0.020$ & $(97.9\%)$ 	& $0.946\pm0.021$ & $(81.8\%)$ & $0.951\pm0.020$ 	& $(98.2\%)$\\
  regular 2 		& $0.995\pm0.003$ & $(99.6\%)$ 	& $0.993\pm0.003$ & $(95.0\%)$ 	& $0.994\pm0.004$ & $(72.9\%)$ & $0.993\pm0.004$ 	& $(94.7\%)$\\
  irregular 1 	& $0.999\pm0.000$ & $(99.9\%)$ 	& $0.999\pm0.001$ & $(96.1\%)$ 	& $0.999\pm0.000$ & $(79.3\%)$ & $0.999\pm0.001$ 	& $(95.6\%)$\\
  irregular 2 	& $0.999\pm0.000$ & $(99.0\%)$ 	& $0.999\pm0.001$ & $(95.0\%)$ 	& $0.999\pm0.001$ & $(75.3\%)$ & $0.999\pm0.001$ 	& $(95.1\%)$\\
  \bottomrule
\end{tabular}%
}
\end{table*}

We have conducted statistical tests to compare the quality of the adversarial and unrecognizable images formed on the closed and open set DNNs.
We use the computed PASS when iterative LOTS reached the target, $\textup{PASS}=0$ when iterative LOTS could not form a perturbation yielding the targeted classification, and $\textup{PASS}=1$ when the original class is the target
(e.g., Fig.~\ref{fig:org}\subref{org:i} is labeled as \textit{coral reef} by Softmax, thus that target is reached).
Two-sided paired t-tests on PASS values show a significant difference $(p<0.00001)$ between the images formed on
Softmax and Openmax when the class aiming vectors (CAVs) are utilized.
When the mean activation vectors (MAVs) are targeted with iterative LOTS, the difference is not statistically significant $(p=0.741)$.
The latter result indicates that with the mean activation vector (MAV) targets the open set DNN is as susceptible to adversarial and unrecognizable examples formed by iterative LOTS as the closed set network.

Finally, to be able to visually compare the examples formed by the four attack scenarios, adversarial and regular unrecognizable images and their perturbations are shown in Fig.~\ref{fig:types}.
We can observe that iterative LOTS targeting the class aiming vector (CAV) or the mean activation vector (MAV) of the \textit{goldfish} class forms similar perturbations on closed and open set DNNs.
However, those similar perturbations yield different prediction probabilities.

%% file: conclusion.tex
\section{Conclusion}

Adversarial example generation techniques mainly focus on forming examples on end-to-end classification networks.
In this paper, we have introduced our novel logits optimized targeting system (LOTS) to directly manipulate the deep features of the penultimate layer in order to form adversarial and unrecognizable images on closed and open set deep neural networks (DNNs).
LOTS can be efficiently used iteratively to form examples on both DNNs.
We have experimentally demonstrated the capabilities of iterative LOTS by generating high quality adversarial and unrecognizable images that are classified as their targeted classes.

We have conducted experiments to compare the robustness of closed and open set DNNs to adversarial and unrecognizable images.
We have generated adversarial and unrecognizable examples on the closed set DNN with the Softmax layer and on the open set DNN utilizing Openmax with two different approaches to target classes.
First, we have used iterative LOTS on activation vectors of the penultimate layer to mimic the specified class aiming vectors (CAVs) and, second, to get closer to mean activation vectors (MAVs) of classes.

To assess the robustness of these DNNs, we have quantified the quality of the produced images using the perceptual adversarial similarity score (PASS), and we have measured the percentage of successful attempts when the formed images were classified as intended.
A less vulnerable system allows lower success rates and/or requires more visible perturbations.
Based on the collected metrics, we have concluded that the open set DNN with Openmax is more robust to adversarial and unrecognizable images when iterative LOTS is used with class aiming vector (CAV) targets.
However, when the mean activation vectors (MAVs) of classes are targeted, DNNs with Softmax and Openmax are equally vulnerable.
While adversarial and fooling images formed on the Softmax layer can be corrected or detected by Openmax \cite{bendale2016towards}, we have shown a novel and efficient attack on Openmax.

Although we have performed our experiments using the mean activation vectors (MAV) that Openmax relies on for class representations, we conjecture that very similar results can be obtained by taking the mean activation vectors of several, arbitrarily selected, and correctly classified examples and using those as targets. Finally, iterative LOTS can be easily extended to work on deep feature representations of any layer in DNNs in order to mimic an arbitrary target representation.